\documentclass[twoside,twocolumn,10pt]{article}

\usepackage{wscg}           
\RequirePackage{ifpdf}
\ifpdf
 \RequirePackage[pdftex]{graphicx}
 \RequirePackage[pdftex]{color}
\else
 \RequirePackage[dvips,draft]{graphicx}
 \RequirePackage[dvips]{color}
\fi
\usepackage[intlimits]{amsmath} 

\usepackage[swedish,english]{babel}
\usepackage{cite}
\usepackage{amsmath,amssymb,amsfonts}
\usepackage{algorithmic}
\usepackage{subcaption}
\usepackage{graphicx}
\usepackage{textcomp}
\usepackage{tabularx,booktabs}
\usepackage{xcolor}
\usepackage{lipsum}  
\usepackage{hyperref}
\usepackage{cleveref}
\def\BibTeX{{\rm B\kern-.05em{\sc i\kern-.025em b}\kern-.08em
    T\kern-.1667em\lower.7ex\hbox{E}\kern-.125emX}}

\newcolumntype{C}{>{\centering\arraybackslash}X} 
\usepackage[switch]{lineno}

\title{Bias mitigation techniques in image classification: fair machine learning in human heritage collections}

\author{
\parbox{0.4\textwidth}{\centering
\textbf{Dalia Ortiz Pablo}\\[1mm]
Centre for Digital Humanities Uppsala\\
Department of ALM\\
Uppsala University\\
Sweden 75126, Uppsala\\[1mm]
dalia.ortiz\_pablo@abm.uu.se
}
\parbox{0.7\textwidth}{\centering
\textbf{Sushruth Badri} \; \; \textbf{Erik Nor{\'e}n} \; \; \textbf{Christoph N{\"o}tzli}\\[1mm]
Department of Information Techology\\
Uppsala University\\
Sweden 75126, Uppsala\\[1mm]
\{sushruth.badri.6580 | erik.noren.3194 | christoph.notzli.7006\}@student.uu.se
}
}

\usepackage{url}
\makeatletter
\def\Uslash{\mathbin{\mathchar`\/}\@ifnextchar{/}{\kern-.15em}{}}
\g@addto@macro\UrlSpecials{\do \/ {\Uslash}}
\def\Ucolon{\mathbin{\mathchar`:}\@ifnextchar{/}{\kern-.1em}{}}
\g@addto@macro\UrlSpecials{\do : {\Ucolon}}
\makeatother

\usepackage{textcomp}

\begin{document}

\twocolumn[{\csname @twocolumnfalse\endcsname

\maketitle  

\begin{abstract}
\noindent

A major problem with using automated classification systems is that if they are not engineered correctly and with fairness considerations, they could be detrimental to certain populations. Furthermore, while engineers have developed cutting-edge technologies for image classification, there is still a gap in the application of these models in human heritage collections, where data sets usually consist of low-quality pictures of people with diverse ethnicity, gender, and age. In this work, we evaluate three bias mitigation techniques using two state-of-the-art neural networks, Xception and EfficientNet, for gender classification. Moreover, we explore the use of transfer learning using a fair data set to overcome the training data scarcity. We evaluated the effectiveness of the bias mitigation pipeline on a cultural heritage collection of photographs from the 19th and 20th centuries, and we used the FairFace data set for the transfer learning experiments. After the evaluation, we found that transfer learning is a good technique that allows better performance when working with a small data set. Moreover, the fairest classifier was found to be accomplished using transfer learning, threshold change, re-weighting and image augmentation as bias mitigation methods.\\

\end{abstract}

\subsection*{Keywords}
Image classification, Fairness, Bias mitigation, Gender classification, Transfer learning, Human Heritage Collection. 

\vspace*{1.0\baselineskip}
}]


\section{Introduction}

\copyrightspace


Artificial intelligence (AI) systems have become a key instrument in many human decision processes. Their presence ranges from basic day-to-day tasks such as listening to music at home using technologies the size of a donut~\cite{mcl19} to transcribing hand-written characters into machine-actionable text data~\cite{cor20}. Those systems present clear benefits, mostly thanks to computers being able to perform tasks at a velocity that humans cannot and without getting tired. However, not everything is positive with AI. One major issue that those algorithms have presented is bias and unfairness. For example, the recruiting algorithm developed by Amazon, where the system learnt key traits from successful applicants'\ resumes to rate and find the top new candidate'\ CVs, exhibited a preference for males in technical positions~\cite{kod19}. This problem does not only occur in contemporaneous sets but also historical scenarios. 

Galleries, archives and museums carry deep insights into human memory and expression. Therefore not only collecting the remains of the past is relevant, but also analysing the objects that have been obtained. Furthermore, it is possible to apply and evaluate the technologies of the present, such as image classification, on the remnants of the past to understand how we can have more nuanced and meaningful descriptions of heritage collections. Moreover, exploring how to connect qualitative issues of human nature, such as bias, diversity and ethics, and the qualitative nature of AI is also important. This work aims to explore the applicability of current deep learning models in image classifications on a collection of pictures of the past, whose major challenges are their diversity as in time, quality and definitions of gender.

Literature on bias and fairness mentions that bias is often encoded in the training data set and that this type of bias affects especially the models' accuracy~\cite{par18}. However, even with well-balanced data sets, bias can be introduced in other steps of the ML pipeline. The work of Wang et al.~\cite{wan19} in gender classification is a good illustration of this. The authors showed that even when their data sets were perfectly balanced, the trained models resulted in biased predictions. Thus, assuming that the input data set is the only source of bias is erroneous. In other studies, researchers have identified, named and classified many other sources of bias, for example, algorithmic bias, which appears due to inappropriate algorithmic choices, and is not present in the input data, or evaluation bias, which happens when the evaluation data set is not representative of the problem or the evaluation metric is incorrect for the task~\cite{fah21, vang22, meh21}. 

In this work, we investigate and evaluate the current techniques for bias mitigation for image classification. The classification task is gender classification using images from cultural heritage collections. Within the scope of this project, gender refers to male or female. The existence of further nuance concerning gender in cultures of the past or today is not within the scope of the project. Further, the data set is only labelled with gender, i.e. the analysis is restricted to gender bias.

\section{RELATED WORK}

\subsection{Bias mitigation}

Technical bias mitigation techniques can be divided into five stages: problem understanding, pre-processing, in-processing, post-processing and deployment \cite{hort22, bell18}. Bias mitigation methods performed on the training data are considered pre-processing; techniques performed while training ML models are categorised as in-processing; and post-processing methods are applied to trained ML models \cite{hort22, bell18}. 

One of the most common techniques for bias mitigation in the pre-processing step is data augmentation - a technique used to increase the amount of data in the training set by performing modifications (e.g. rotations, colour transformation, reflection ) on the original set~\cite{mij18}. In the work done by McLaughlin et al.~\cite{mcl15}, the authors evaluated the effectiveness of different data augmentation techniques in re-identification systems. They found that changing an image background increases the performance of a model only when a combination of other augmentation techniques, such as cropping and mirroring, are also used.

Other bias mitigation methods have also been investigated in the literature. The work of Wang et al. \cite{wan20} evaluates strategic re-sampling, adversarial training, domain discriminative training, and domain-independent training in a gender classification scenario using CNNs and a data set composed of pictures of celebrities. The authors found that oversampling outperformed the other techniques, and that was followed closely by domain-independent training. Similarly, Lee et al. \cite{lee22} revise bias mitigation methods for CNNs, e.g. ReBias and vanilla, to create a benchmark for the bias mitigation pipeline. The study showed that state-of-the-art approaches achieved different approaches depending on the training data set, which suggests that a bias mitigation process is task specific.



\subsection{The FairFace data set}\label{sec:ff}

Neural network models have been shown to learn and amplify biases in training data \cite{hal22}. This is partly the motivation for the creation of the FairFace data set, presented in the work by K{\"a}rkk{\"a}inen and Joo \cite{kar19}. The FairFace data set contains 108 501 images, and it is balanced concerning gender, ethnicity, and age, and therefore does not suffer from the same bias that other big data sets do. In \cite{kar19}, it is shown that models trained with the FairFace data set generalize better than other existing face data sets to unseen and ethnically diverse data. In the work of Kotti et al. \cite{kot22}, the data set is used in their experiments for evaluating bias in Generative Adversarial Networks (GANs).  Moreover, a model trained on the FairFace data set was used in \cite{dev22} in order to benchmark the performance of their new fair model. In our work, we conduct transfer learning using the FairFace data set for the models to learn first for a known fair data set. 

\section{BACKGROUND}\label{sec:back}

\subsection{Deep learning models}\label{sec:model}

Our classifiers are based on existing networks provided in the tensorflow framework. We chose Xception \cite{chol17} and EfficientNet \cite{tan19} as base networks. Both were implemented using Keras' model API. 

The Xception network was first introduced in 2017 in the paper "Xception: Deep Learning with Depthwise Separable Convolutions" by Chollet \cite{chol17}. The model is a fully convolutional network designed with the goal of being a more efficient variant of the Inception architecture from 2014 \cite{sze15}. The technique that separates the Xception architecture from the Inception architecture is the implementation of depthwise separable convolutions instead of regular convolutions. Regular convolutions work by performing convolutions on all channels at once, unlike depthwise separable convolutions which performs a single convolution operation on each input channel. The Xception architecture has 36 convolutional layers and has an input size of 299x299. 

EfficientNet was introduced in the 2019 paper "EfficientNet: Rethinking Model Scaling for Convolutional Neural Networks" \cite{tan19}. Rather than being a single network, EfficientNet is best described as a family of networks architectures, created from scaling the baseline network EfficientNet-B0. The paper introduces a new approach to scaling models which included scaling the network's width, depth and image resolution together. The EfficientNet architecture that we implemented was EfficientNet-B3 which has an input size of 300x300 which is similar to Xception. In \cite{tan19}, the EfficientNets' performances, compared to other convolutional neural networks, is the state-of-the-art of several data sets, while reducing the size of the models. The EfficientNet-B3 architecture is about half the size of Xception, with 12 million versus 23 million parameters. 

Both the models have the same four top layers appended to produce the binary classification. These consists of a GlobalAveragePooling2D layer, a BatchNormalization layer, a Dropout layer with a dropout rate of 0.2, and lastly, a dense layer for binary classification with a sigmoid activation function~\cite{ker15}.

\subsection{Fairness metrics}\label{sec:fairness_metrics}

There exist several criteria to quantify bias. Different fairness metrics are built upon different definitions of fairness, and therefore, give priority to different aspects of classification performance. In a survey of 341 publications relating to bias mitigation by Garg et al., the relationship between fairness metrics and bias mitigation methods was explored \cite{gar20}. According to the survey, the two most common categories of fairness metrics are \textit{Definitions Based on Predicted Outcome} and \textit{Definitions Based on Predicted and Actual Outcome}, the survey adopts these which were introduced in \cite{ver18}. Some examples of fairness metrics that belong to these categories are:

\begin{itemize}
    \item  Demographic Parity Difference (Statistical Parity Difference) - Ideally the same amount of positives across classes. Definition is based on Predicted Outcome~\cite{ver18}. 
    \begin{equation}
        \text{positives}_{0} = \text{positives}_{1}
    \end{equation}
    With positives
    \begin{equation}
        \text{positives} = \text{TP} + \text{FP}
    \end{equation}
    In the binary case:
    \begin{equation}
        \text{TP} + \text{FP} = \text{TN} + \text{FN}
    \end{equation}

    \item Proportional Parity Difference - Ideally the same amount of normalized positives across classes~\cite{koz21}. 
    \begin{equation}
        \frac{\text{positives}_{0}}{\text{all}} = \frac{\text{positives}_{1}}{\text{all}}
    \end{equation}
    With all
    \begin{equation}
        \text{all} = \text{TP} + \text{FP} + \text{TN} + \text{FN}
    \end{equation}
    In the binary case:
    \begin{equation}
        \frac{\text{TP} + \text{FP}}{\text{TP} + \text{FP} + \text{TN} + \text{FN}} = \frac{\text{TN} + \text{FN}}{\text{TP} + \text{FP} + \text{TN} + \text{FN}}
    \end{equation}
    
    \item Equality of Opportunity - Ideally the same true positive rate across classes. Definitions are based on Predicted and Actual Outcome~\cite{gar20}.
    \begin{equation}
        \text{TPR}_{0} = \text{TPR}_{1}
    \end{equation}  
    In the binary case:
    \begin{equation}
        \frac{\text{TP}}{\text{TP} + \text{FN}} = \frac{\text{TN}}{\text{TN} + \text{FP}} 
    \end{equation}

    \item Predictive Rate Parity Difference - Ideally the same positive predictive value across classes~\cite{cho17}. Definitions are based on Predicted and Actual Outcome \cite{ver18}.
    \begin{equation}
        \text{PPV}_{0} = \text{PPV}_{1}
    \end{equation}  
    With PPV
    \begin{equation}
        \text{PPV} = \frac{\text{TP}}{\text{TP} + \text{FP}}.
    \end{equation}  
    In the binary case:
    \begin{equation}
        \frac{\text{TP}}{\text{TP} + \text{FP}} = \frac{\text{TN}}{\text{TN} + \text{FN}} 
    \end{equation}

\end{itemize}

The scope of the survey by Garg et al.~\cite{gar20} covers fairness metrics for ML models in general, but the fairness metrics are nonetheless applicable in the deep image classification use case as well. Examples  being Demographic Parity being used in \cite{fra21} and Equality of Opportunity being used in \cite{sto22, zha18}. Therefore, in project, we use demographic parity difference, equality of opportunity, proportional parity difference and predictive rate parity difference as fairness metrics.  



\subsection{Bias mitigation techniques}\label{sec:bias}

\subsubsection{Reweighting} Reweighting is a pre-processing bias mitigation method. The idea of this method is to give classes that are more common in the training data set a lower weight i.e. the sample has less effect on the training of the model. Reweighing approach also maintains a high accuracy level \cite{kam12}. Pre-processing techniques try to transform the data so that the underlying discrimination is removed \cite{meh21}. The class weights are assigned and passed to the model while training through Keras implementation.

\subsubsection{Image augmentation} Image augmentation is an in-processing method for bias mitigation. In this project, data augmentation is implemented by altering the training data in order to deal with classification bias in under-representation of certain groups. Data augmentation can reduce classification error for discriminated groups. Furthermore, even though different classifiers do not perform equally good, they exhibit positive results when data augmentation takes place \cite{ios18}. In our project the data augmentation has four pre-processing layers that are added at the beginning of the model. These layers perform random flip, random rotation, random translation (i.e. random movement), and random contrast. The pre-processing layers are implemented using the Keras API for image augmentation layers \cite{ker15}. 

\subsubsection{Threshold change} Changing the threshold of the model is a post-processing method \cite{hort22}. In the standard implementation of our model we use a threshold of 0.5 i.e. every predicted value below 0.5 is interpreted as Female. This threshold is not necessarily optimized for fairness. We applied the following threshold changes to the models:

\textbf{Equal true} In this case we optimize the threshold to the minimal difference of predicted true values in each class:
\begin{equation}
    \min\{\lvert \text{TP}_{t}-\text{TN}_{t}\rvert\} \quad \text{with t} = [0,1]
\end{equation}
\textbf{Equal false} In this case we optimize the threshold to the minimal difference of predicted false values in each class.
\begin{equation}
    \min\{\lvert \text{FP}_{t}-\text{FN}_{t}\lvert\} \quad \text{with t} = [0,1]
\end{equation}
\textbf{Equal total} In this case we optimize the threshold to predict the minimal difference of predicted values in each class.
\begin{equation}
    \min\{\lvert(\text{TP}_{t}+\text{FP}_{t})-(\text{TN}_{t}+\text{FN}_{t})\rvert\} \quad \text{with t} = [0,1]
\end{equation}
\textbf{Equal opportunity} In this case we optimize the threshold to predict the minimal difference of predicted values in each class.
\begin{equation}
    \min\{\lvert \frac{\text{TP}_{t}}{\text{TP}_{t} + \text{FN}_{t}} - \frac{\text{TN}_{t}}{\text{TN}_{t} + \text{FP}_{t}}\rvert\} \quad \text{with t} = [0,1]
\end{equation}

\subsubsection{Transfer Learning}
\label{sec:tl}
Transfer learning is the ML technique of taking the knowledge that is able to be learned by training a model on one task and then fine tuning it to a different but related task~\cite{pan10}. In this project, transfer learning was implemented as a way overcome training data scarcity. For this part of the process, the FairFace data set is the source domain and the cultural heritage data set is the target domain. The tasks are similar in both domains, being binary image classification of gender in both cases. 

Through our experiments, it will be investigated if it is possible to transfer a fair model trained on the FairFace data set. This will be done by comparing the results of the models implementing transfer learning against the ones trained on only target domain. The metrics for evaluating bias will be fairness metrics detailed in~\cref{sec:fairness_metrics} and~\cref{sec:bias}.



\subsection{Cultural Heritage Data Set}\label{sec:dataset}

The Cultural Heritage Data Set (CHDS) contains labelled images from the The National Museums of World Culture \cite{VKM} in Sweden, and the data set was manually post-processed as part of an earlier part of the Quantifying Culture project. Within the data set there are 3128 images in total of people, where 1067 images are labeled male, and 2061 labeled female. These images are photographs taken "in-the-wild" and do not guarantee that they contain only one person at a time, but ensure that all the individuals are of the same sex. The time period which the photographs stem from are either the nineteenth or twentieth century. After performing the face extraction described in~\cref{sec:faceextrac}, 4897 faces were detected, 2331 labeled male, and 2566 labeled female. 


\section{Implementation details}

\subsection{Face extraction}\label{sec:faceextrac}
Since the CHDS contains full-body images of people, and our model is designed to perform gender classification based on facial features, face extraction is performed on the CHDS. The face extraction module makes use of the Python API of the dlib ML toolkit \cite{kin09}. Within the module a pre-trained CNN model is used for face detection which then allows for extracting cropped images of the faces contained in the photographs of the CHDS. 

\subsection{Experimental setup}

We use three experimental setups to test our networks. In the first experiment, we used only the CHDS to train and test our networks. In a second step, we used the FairFace data to pretrain validate and test them our models. The third experiments use the FairFace models from the second experiments for transfer learning. The transfer learned models are then validated, tested and compared to the baseline models. For all experiments the fairness metrics discussed in \cref{sec:fairness_metrics} are implemented for evaluation. 

\subsubsection{Baseline CHDS experiments} In the baseline model experiments we train and cross validate the models shown in~\cref{sec:model}. For each of these two models, there are three variations tested, which depend on the bias mitigation method being evaluated. The first variation is without any bias mitigation method;  the second is with class re-weighting applied; and the third is with data set augmentation applied. These bias mitigation methods are covered in~\cref{sec:bias}. The models were trained for a maximum of 20 epochs with the learning rate decreasing exponentially. We added an early stopping to the training, which means that the training will stop in case the training does not decrease the validation loss in three consecutive epochs.

\subsubsection{FairFace experiments} The models described in~\cref{sec:model} are pretrained with the help of the FairFace data in three different ways. First we just train the models, in a second experiment we reweight the classes because there is a slight bias towards the Male class in the FairFace data. In the third experiment we add the augmentation layers described in~\cref{sec:bias}. These experiments are run to create the different variations of base models to be used for transfer learning, as well as a way to evaluate performance on the models on a data set that we know to be fair and balanced. 

\subsubsection{Transfer learning experiments} 


To build some the classifiers for this project, we train a base model, either EfficientNet-B3 or Xception, with the FairFace training data. During training, the weights of the models' layers are tuned and adjusted in order to increase accuracy. This trained model is then moved to the target domain and has a number of its layers and their parameters frozen in order to keep the knowledge learned in the source domain. For transfer learning with the EfficientNet-B3 model all layers except the the last forty layers are frozen. In the Xception case all layers except the last ten layers and the four top layers are frozen. Also, we compare four different version of the transfer learned models. We use the unweighted and unaugmented pretrained model to test transfer learning. 


\subsubsection{Data set splits} As mentioned in~\cref{sec:dataset} the cultural heritage data set does not contain a lot of data. Therefore we choose  to use 80\% of the data for training, 10\% for validation and 10\% for testing. We used the same splits for the training with the help of the FairFace data set. For cross validation we used a 5-Fold cross validation split, 80\% of data for training and 20\% for validation.

\subsubsection{Early stopping} We use early stopping provided by Keras \cite{ker15}. I.e. the training stops when the validation loss is not reduced in three consecutive training steps.

\subsubsection{Used infrastructure for training} For the training of the models we used the Alvis cluster of Chalmers University \cite{alvis}. We used the A100 GPUs of the cluster. This allowed us to use larger batch sizes and smaller training time.  

\section{Results}
\label{sec:results}
The results of our experiments can be found in this section. Within \cref{tab:results} the results of the Baseline CHDS experiments and the Transfer learning experiments can be seen. Positive values in the fairness metrics Demographic Parity Difference, Proportional Parity Difference and Predictive Rate Parity Difference show a bias in favor of the Male class. In the Equality of Opportunity metric the bias is in favor of the Male class if the values are negative.

\begin{table*}
    \scriptsize
    \begin{tabularx}{\textwidth}{@{} l *{10}{C} c @{}}
        \toprule    
        Network name   
        & Transfer learning & Threshold change & Reweighting & Image augmentation & Accuracy 
        & Demographic Parity Difference & Proportional Parity Difference & Equality of Opportunity & Predictive Rate Parity Difference \\ 
        \midrule
        EfficientNet & No & No & No & No & 78.5 +/- 1.2\% &  -39 +/- 54.6 & -0.04 +/- 0.06 & 0.01 +/- 0.06 & -0.03 +/- 0.05 \\
        EfficientNet & No & Equal false & No & No & 78.9\% & 4 & 0.008 & 0.036 & -0.053 \\
        
        EfficientNet & No & No & Yes & No & 80.4 +/- 1.6\% & -64 +/- 33.8 & -0.06 +/- 0.03 & 0.04 +/- 0.03 & -0.01 +/- 0.03 \\
        EfficientNet & No & Equal total & Yes & No & 79.9\% & 6 & 0.012 & 0.041 & -0.056 \\
        
        EfficientNet & No & No & No & Yes & 78 +/- 2.2\% & -155 +/- 77.8 & -0.16 +/- 0.08 & 0.13 +/- 0.09 & 0.04 +/- 0.06  \\
        EfficientNet & No & Equal opp. & No & Yes & 80.1\% & -8 & -0.016 & 0.013 & -0.039 \\
        \addlinespace
        EfficientNet & Yes & No & No & No & 84.5 +/- 0.5\% & -45 +/- 42.3 & 0.05 +/- 0.04 & 0.01 +/- 0.04 & 0.02 +/- 0.03 \\
        EfficientNet & Yes & Equal false & No & No & 83.7\% & -8 & -0.016 & 0.017 & -0.038 \\
        
        EfficientNet & Yes & No & No & Yes & 83.6 +/- 0.3\% & -115 +/- 26.2 & -0.11 +/- 0.02 & -0.08 +/- 0.04 & 0.03 +/- 0.04 \\
        EfficientNet & Yes & Equal false & No & Yes & 82.1\% & 12 & 0.024 & 0.056 & -0.064  \\
        
        EfficientNet & Yes & No & Yes & No & 84.2 +/- 1.5\% & -101 +/- 73.5 & -0.1 +/- 0.07 & 0.07 +/- 0.08 & 0.02 +/- 0.06 \\
        EfficientNet & Yes & Equal false & Yes & No & 80.7\% & 14 & 0.028 & 0.059 & -0.066 \\

        EfficientNet & Yes & No & Yes & Yes & 83.9 +/- 1.4\% & -13.4 +/- 48.2 & -0.01 +/- 0.05 & 0.02 +/- 0.04 & -0.04 +/- 0.03 \\
        EfficientNet & Yes & Equal total & Yes & Yes & 82.1\% & 0 & 0 & 0.031 & -0.049  \\
        \addlinespace
        \hline
        \addlinespace
        Xception & No & No & No & No & 52.4\% & -492 & -1 & -1 & -0.524  \\
        
        Xception & No & No & Yes & No & 47.6\% & 492 & 1 & 1 & 0.476  \\
        
        Xception & No & No & No & Yes & 52.4\% & -492 & -1 & -1 & -0.524  \\
        \addlinespace
        Xception & Yes & No & No & No & 65.6 +/- 3.7\% & -7 +/- 435 & -0.01 +/- 0.44 & -0.01 +/- 0.45 & -0.06 +/- 0.15  \\
        Xception & Yes & Equal total & No & No & 78.3\% & 2 & 0.004 & 0.032 & -0.051 \\

        Xception & Yes & No & No & Yes & 72.4 +/- 0.1\% & -338 +/- 401 & -0.34 +/- 0.40 & -0.33 +/- 0.41 & 0.1 +/- 0.16 \\
        Xception & Yes & Equal opp. & No & Yes & 80.9\% & -16 & -0.033 & -0.002 & -0.029 \\
        
        Xception & Yes & No & Yes & No & 66.4 +/- 6\% & -341 +/- 398 & -0.35 +/- 0.41 & -0.34 +/- 0.41 & 0.09 +/- 0.19 \\
        Xception & Yes & Equal opp. & Yes & No & 77.4\% & -14 & -0.028 & -0.002 & -0.033 \\

        Xception & Yes & No & Yes & Yes & 73.1 +/- 7.8\% & -389 +/- 169 & -0.39 +/- 0.17 & -0.37 +/- 0.19 & 0.15 +/- 0.08 \\
        Xception & Yes & Equal opp. & Yes & Yes & 81.3\% & 4 & 0.008 & 0.039 & -0.054 \\
        \bottomrule
    \end{tabularx}
    \caption{Validation results of the experiments. The results of the CHDS experiments are the entires marked with "No" in the transfer learning column. Entries marked "Yes" in the Transfer learning column are implemented with transfer learning from the FairFace data set. In the table the performance of different combinations of the different bias mitigation methods can be seen.}
    \label{tab:results}
\end{table*}

\subsection{Baseline CHDS experiments}
\begin{figure}[h]
    \centering
    \includegraphics[width=0.48\textwidth]{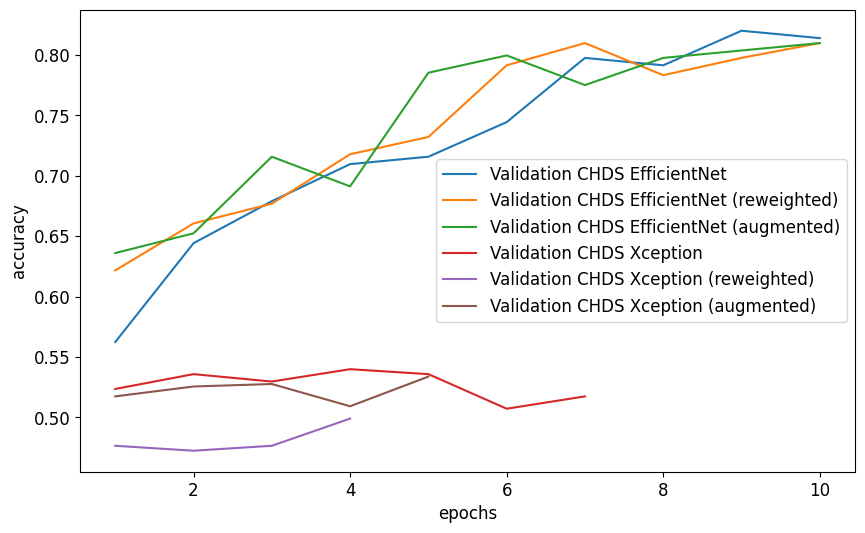}
    \caption{EfficientNet and Xception validation accuracy for the CHDS experiments.}
    \label{fig:acc_comp_chds}
\end{figure}

\begin{figure}[h]
    \centering
    \includegraphics[width=0.48\textwidth]{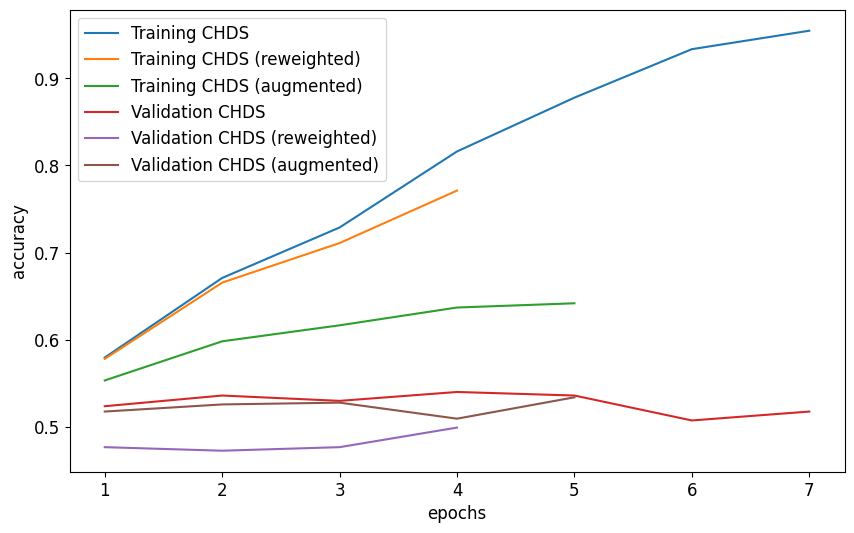}
    \caption{Xception training and validation accuracy for the CHDS experiments.}
    \label{fig:acc_xcp_chds}
\end{figure}

Baseline CHDS experiments were performed for both the EfficientNet-B3 and Xception models. Figure \ref{fig:acc_comp_chds} shows the performance in accuracy for EfficientNet and Xception. It can be seen that all three variations of the EfficientNet model reaches higher accuracy than the Xception variations. Figure  \ref{fig:acc_xcp_chds} shows the training and validation accuracies for the Xception variations. The EfficientNet variations also have greater performance in the fairness metrics as well as seen in~\cref{tab:results}.

\subsection{FairFace experiments}
\begin{figure}
    \centering
    \includegraphics[width=0.48\textwidth]{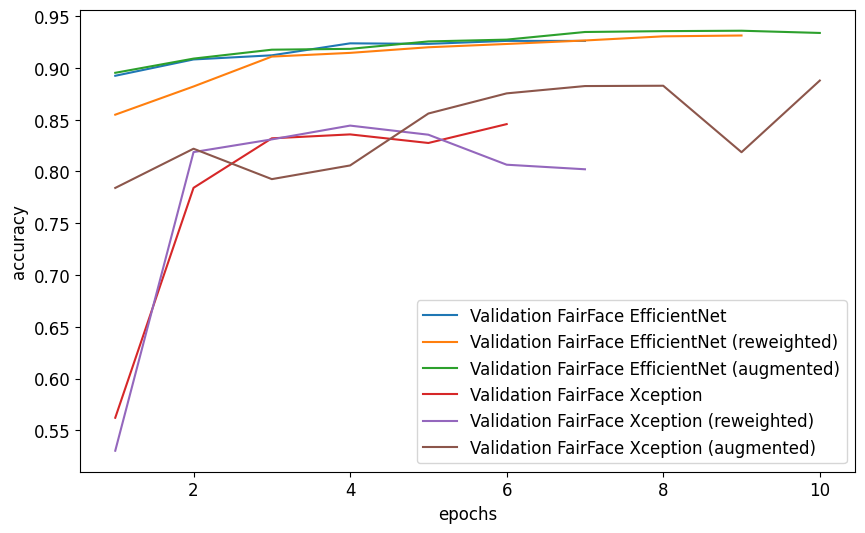}
    \caption{EfficientNet and Xception validation accuracy for the FairFace experiments.}
    \label{fig:acc_comp_ff}
\end{figure}

\begin{figure}
    \centering
    \includegraphics[width=0.48\textwidth]{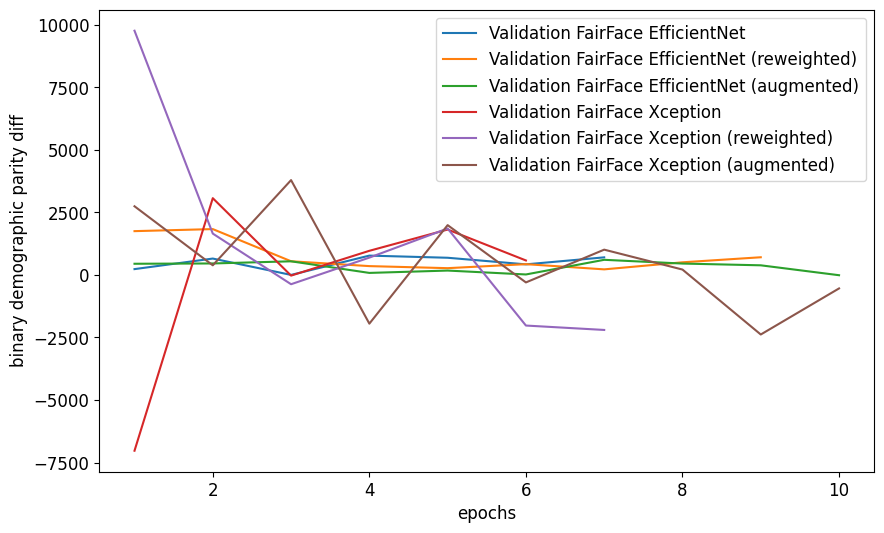}
    \caption{EfficientNet and Xception validation Demographic Parity Difference performance for the FairFace experiments.}
    \label{fig:dem_par_diff_comp_ff}
\end{figure}

\begin{figure}
    \centering
    \includegraphics[width=0.48\textwidth]{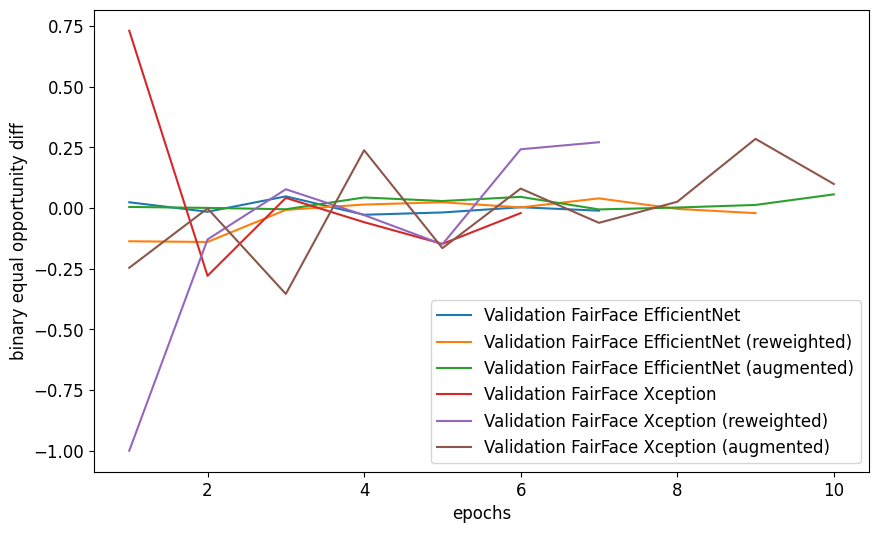}
    \caption{EfficientNet and Xception validation Equality of Opportunity performance for the FairFace experiments.}
    \label{fig:eq_opp_diff_comp_ff}
\end{figure}

The model that achieves the highest performance in accuracy on the FairFace data set is EfficientNet with a validation accuracy of $\sim$91\% as seen in Figure \ref{fig:acc_comp_ff}. The best performing Xception model is the one implemented with augmentation which achieves a validation accuracy of $\sim$87\%. Figure \ref{fig:dem_par_diff_comp_ff} shows the performance of the models with regards to Demographic Parity Difference where all variations except Xception with reweighting have similar results. This is also the case when evaluating with regards to Equality of Opportunity Difference, as seen in Figure \ref{fig:eq_opp_diff_comp_ff}. 

\subsection{Transfer learning experiments}
Using the models created in the FairFace experiments for transfer learning, the accuracy results shown in \cref{tab:results} were achieved. EfficientNet achieves the highest accuracy of the transfer learned models with 83.7\%. In the fairness metrics different transfer learning variation of both EfficientNet and Xception have the best performance. Measured in Demographic Parity Difference the EfficientNet model with equal total threshold change, re-weighting, and image augmentation performs best. If Equality of Opportunity is considered instead the re-weighted Xception model is best performer. When using augmentation and the EfficientNet network the bias metrics stay close to 0 during the training as seen in \Cref{fig:comp_pred_transfer} and \Cref{fig:comp_pro_transfer}.

\section{Discussion}\label{sec:discussion}

\begin{figure}
    \centering
    \includegraphics[width=0.48\textwidth]{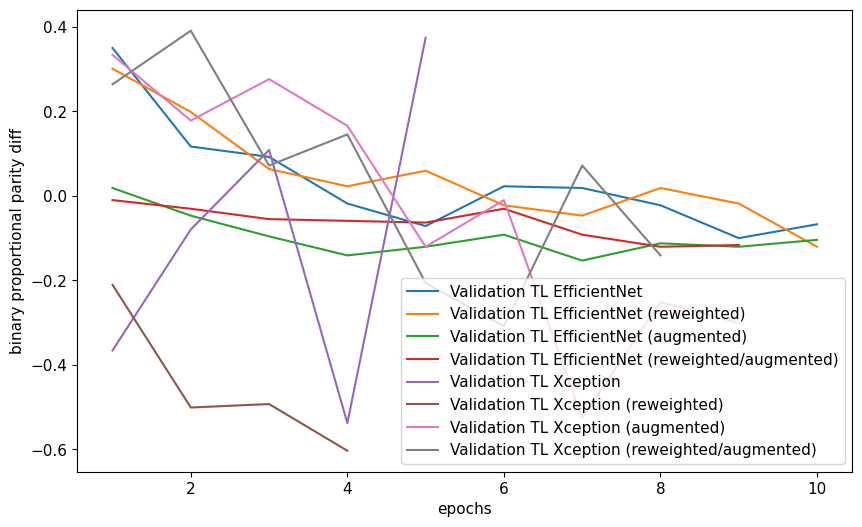}
    \caption{EfficientNet and Xception validation Proportional Parity Difference for the transfer learning experiments.}
    \label{fig:comp_pro_transfer}
\end{figure}

\begin{figure}
    \centering
    \includegraphics[width=0.48\textwidth]{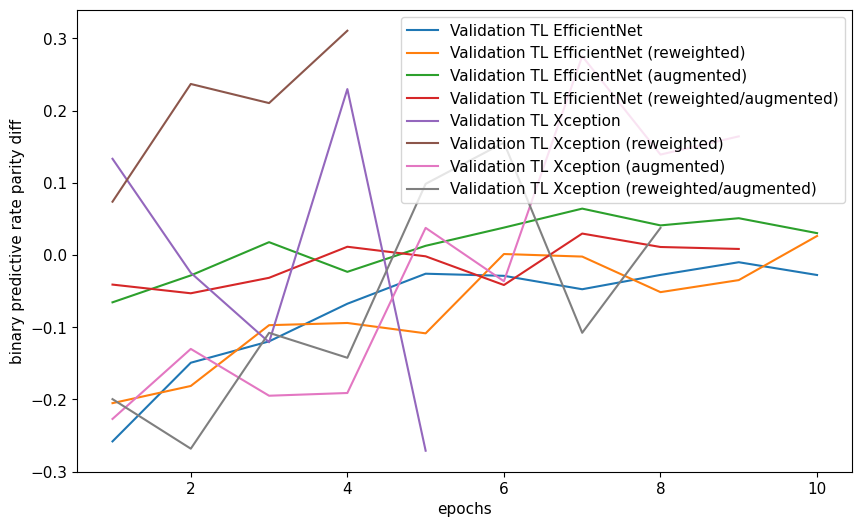}
    \caption{Xception training and validation Predictive Rate Parity Difference for the transfer learning experiments.}
    \label{fig:comp_pred_transfer}
\end{figure}

The Xception network is not ideal for a use case with a small data set, in this case the CHDS. In \cref{tab:results} the results show that the Xception network is not performing well due to overfitting as seen in Figure \ref{fig:acc_xcp_chds}. More regularization within the network and the final layers could be a possible solution to fix the overfitting in the base case of the Xception model. However, here the value of transfer learning is shown to be an effective way to improve a model which poor performance is partly due to lack of training data. Xception used with transfer learning allowed for a validation accuracy of 73.1\%, which greatly outperforms the baseline CHDS trained models as seen in Table \cref{tab:results}.  

The results in \cref{sec:results} show that EfficientNet has a higher accuracy and is better than Xception in regard to fairness in the transfer learning variations. These results are expected since EfficientNet is a newer network and also shows better performance in the benchmarks \cite{tan19}. 

Using transfer learning to train the models is a good choice in our use case. The biggest advantage is that we get a higher accuracy due to training the models with more data. Comparing the EfficientNet model, with or without transfer learning, it can be seen in \cref{tab:results} that accuracy in increased from 78.5\% to 84.5\%. However, as a method for bias mitigation transfer learning does not make much of an impact. For the experiments on the EfficientNet variations the fairness metrics do not appear to be influenced by applying transfer learning or not. Due to the poor performance of the baseline variations of Xception, it is not possible to see if previous results translate to the Xception model.

When transfer learning and augmentation are used together with reweighting on EfficientNet, the fairness metrics are improved. This is also the case for accuracy as seen in \cref{tab:results}. Using reweighting alone does not show any significant improvement in the metrics for both EfficientNet and Xception. It could be due to the fact that there is not a big difference in the amount of samples in each of the classes in CHDS.

Comparing the EfficientNet CHDS results in~\cref{tab:results}, with and without image augmentation, it can be seen that accuracy is not changed much, the accuracy score being 78.5\% +/- 1.2\% without and 78\% +/- 2.2\% with. Moreover, when evaluating these same two variations by the fairness metrics no apparent improvement can be seen in any case. This pattern emerges in the transfer learning experiments as well for EfficientNet. Using image augmentation alone slightly worsens performance in terms of accuracy and improvement cannot be seen in the fairness metrics either. However, when comparing the transfer learning variations, with and without image augmentation, with reweighting implemented, the pattern does not repeat. In this case image augmentation improved performance in all accord with all fairness metrics while reaching a perfect score for Demographic- and Proportional Parity. For transfer learning experiments for Xception, image augmentation improves the fairness metrics when paired together with reweighting, similarly to EfficientNet. Image augmentation improves accuracy for the Xception variations. 
One important observation is that augmentation stabilizes the training (metrics always close to the ideal value) for EfficientNet as seen in \Cref{fig:comp_pred_transfer} and \Cref{fig:comp_pro_transfer}. This is advantageous because we can stop the training at any point with similar bias metric values.

When applied, threshold change improves performance in the fairness metrics consistently. In all EfficentNet experiments the improvement in fairness came at the cost of accuracy, this is however not translated in the Xception experiments. It could be because of Xception network overfitting on the data and it favours a certain class during prediction. This can be seen from the metric values in \ref{tab:results}. When using threshold change, some of the wrong predictions move over to the other side of the decision boundary and so accuracy improves.  

Regarding the CHDS data set, it shows a slight bias favouring the Female class (52.4\% of the images in the Female class). This is also the case for our overall best model (EfficientNet, with reweighting and image augmentation but no thresholding). For example, the value of bias according to the Proportional Parity Difference is 1 +/- 5\% towards the Female class. Gender is something that should be considered concerning fairness; however, there is a cause to consider fairness in relation to other metrics as well. With the CHDS data set only having gender as its label, further investigation into other sources of bias is difficult to attain. The CHDS data set is diverse concerning age, ethnicity, culture, and the period the photo was taken. All of these aspects add the possibility for bias. Photographs portraying a given group of people of a certain ethnicity could have been taken in poorer conditions than photographs of another group due to the time period or other external conditions. As a result, this could lead to it being harder for the model to learn to classify the first group correctly, which might not show up in our results evaluated by gender.

Due to using a pre-trained model in the face extraction module for the cultural heritage data set (\cref{sec:dataset}), there is an additional potential source of bias. For example, it is possible that the pre-trained model was itself trained with biased data, thus leading to it potentially having a better ability to extract the faces of a certain group. Since we did not investigate the total amount of faces belonging to each class in the photographs in the CHDS, no evaluation of the potential bias of the face extraction module was performed. 

\section{Conclusion}\label{sec:conclusion}

In this paper, we evaluated three novel bias mitigation techniques for image classification in cultural heritage data sets. We found that individually implementing augmentation, class re-weighting, and threshold change does not lead to a fairer model compared to the baseline model. We have also evaluated the performance of the classifiers when implementing transfer learning. We have shown that, for this task, combining transfer learning with image augmentation, class re-weighting, and threshold change is the best way to reach a fair classifier.

\section{Future work}
In our current implementation the process of fine tuning was not included in the transfer learning implementation. In fine tuning the final models are trained, with all layers unfrozen, with a very low learning rate for just a few epochs. This could potentially improve performance, mainly accuracy. 

Further work could be done in tuning the hyper-parameters of our models. Since the priority of this project was the evaluation of the bias mitigation methods, less focus was put on perfecting model performance. Therefore work can be done in investigating optimal learning rate, batch size, weight decay etcetera. 

As discussed in~\cref{sec:discussion}, improvement to our analysis could be done if we would have had access to multiple labels, e.g. age, ethnicity, etcetera. With more labels we could take a look a the bias from an intersectional stand point. With the help of intersectionality we could show further biases in the approach, and we could take action to reduce these biases.

Another interesting addition to the project would be to compare the results of the EfficentNet with an implementation of the EfficientNetV2. EfficientNetV2 includes data augmentation layers and we had promising preliminary results for the base model.

\section{Acknowledgments}

This work has been partially supported by the WASP-HS (Autonomous Systems and Software Program-Humanities and Society) grant - an initiative of the Wallenberg Foundations; project title: "Quantifying Culture: AI and Heritage Collections"; project number: MAW 2020.0054. The computations/data handling were enabled by resources provided by the Swedish National Infrastructure for Computing (SNIC) at Chalmers Centre for Computational Science and Engineering (C3SE) partially funded by the Swedish Research Council through grant agreement no. 2018-05973, project Dnr: SNIC 2022/22-1091.

%
%




\end{document}